\documentclass{article}

\usepackage{natbib}
\setcitestyle{authoryear,open={(},close={)}} 

\usepackage{url}
\usepackage{graphicx}
\usepackage{booktabs}
\usepackage{multirow}
\usepackage{float}
\usepackage{csquotes}
\usepackage{multirow, array}
\usepackage{amsmath, amssymb, booktabs}
\usepackage{amsmath}
\usepackage{tikz}
\usepackage{float}
\usepackage{caption}
\usepackage{framed}
\usepackage[hidelinks]{hyperref}

\date{}

\makeatletter
\define@key{Gin}{side-by-side}[]{\setkeys{Gin}{width=0.475\linewidth}}
\makeatother

\usetikzlibrary{shapes,decorations,arrows,calc,arrows.meta,fit,positioning}
\tikzset{
    -Latex,auto,node distance =1 cm and 1 cm,semithick,
    state/.style ={ellipse, draw, minimum width = 0.7 cm},
    point/.style = {circle, draw, inner sep=0.04cm,fill,node contents={}},
    bidirected/.style={Latex-Latex,dashed},
    el/.style = {inner sep=2pt, align=left, sloped}
}

\title{Identifying attributions of causality in political text}

\author{Paulina García-Corral\footnote{Hertie School, corral@hertie-school.org}}

\begin{document}

\maketitle

\begin{abstract}
Explanations are a fundamental element of how people make sense of the political world. Citizens routinely ask and answer questions about why events happen, who is responsible, and what could or should be done differently. Yet despite their importance, explanations remain an underdeveloped object of systematic analysis in political science, and existing approaches are fragmented and often issue-specific. I introduce a framework for detecting and parsing explanations in political text. To do this, I train a lightweight causal language model that returns a structured data set of causal claims in the form of cause-effect pairs for downstream analysis. I demonstrate how causal explanations can be studied at scale, and show the method's modest annotation requirements, generalizability, and accuracy relative to human coding. 
\end{abstract}

\section{Introduction}
\label{sec:introduction}

Causal attributions are claims that link an outcome to a cause \citep{kirfel_inference_2022}. Causality is so embedded in human reasoning that causal attributions have been shown to emerge immediately in times of crisis \citep{graham_outbreak_2024}, as well as offered spontaneously when people are asked to think about political issues \citep{Iyengar1987}. Furthermore, because causal attributions are relational, rather than treating actors and events as isolated, they highlight the underlying relational reasoning people use to connect events, assign responsibility, and justify actions \citep{vossing_argument-stretching_2023}.

Framing is fundamentally a process of making causal explanations, or communicating causal attributions:

\begin{displayquote}
``[Frames] \textit{define problems}--determine what a \textbf{causal} agent is doing with what costs and benefits, usually measured in terms of common cultural values; \textit{diagnose \textbf{causes}}--identify the forces creating the problem; \textit{make moral judgments}--evaluate \textbf{causal} agents and their \textbf{effects}; and \textit{suggest remedies}--offer and justify treatments for the problems and predict their likely \textbf{effects}.''\citep[p. 52, italics in original, boldface mine for emphasis]{Entman_1993}
\end{displayquote}

Political actors engage in framing, and routinely offer causal explanations and counter explanations for the same events by assigning blame, claiming credit, or justifying policy decisions that favor them or their party \citep{robison_can_2022}. Explanations are also shaped by partisan and in-group bias: citizens tend to minimize their preferred party’s responsibility for negative outcomes while amplifying its role in positive ones \citep{rudolph2006, Jefferson_Neuner_Pasek_2021}. 

Furthermore, explanations have important downstream effects on public opinion, and on political attitudes and behavior. For instance, causal explanations in the media shape political attitudes. Citizens’ interpretations of political events are directly influenced by media framing, which in turn affects their evaluations of presidential performance \citep{Iyengar1987}. Similarly, legislators’ explanations can boost support from both partisans and nonpartisans \citep{Grose2015}. Elite explanation-giving is also positively associated with satisfaction with democracy and government among voters \citep{robison_explanation_2024}. Taken together, literature from across political science highlights the causal dimension of political thinking and its effects: causal explanations are not neutral expressions of causality, but rhetorical tools used to persuade, justify, and mobilize. Even when misaligned with reality, they shape how citizens interpret complex events and, ultimately, how they act in the democratic process.

However, despite the centrality in political reasoning and public discourse, causal attributions have been difficult to systematically study at scale, and research that engages with the topic remains fragmented, often lacking a unified perspective on its central function. Previous approaches to study attribution often rely on domain specific coding schemes or limited survey experiments. For example, for content analysis, researchers have used individual vs. societal attributions \citep{sotirovic_how_2003}, national vs. foreign \citep{doi:10.1086/703208}, blame vs. claim \citep{heinkelmann2020}, national vs. regional \citep{Tilley2011}, or us vs. them \citep{hameleers_they_2017}. While in survey experiments, designs tend to focus on a single actor or domain, testing different framing strategies such as adding a politician's job title and party affiliation \citep{Malhotra2008} or including competence cues \citep{graham_outbreak_2024}. These different endeavors have limited the ability to understand the broader structure of causal attribution in the public sphere. To examine explanations as a whole, a general, scalable method that can detect and analyze causal claims is needed, such that we can easily answer the question ``\textit{who is claiming what and about whom?}'' across different contexts, with minimal manual intervention.

\section{Building structured representations of causal attributions at scale}
\label{sec:Building_structured_representations_of_causal_attributions}

Causal attributions usually follow the pattern ``\textbf{E} happened because of \textbf{C}'' \citep{kirfel_inference_2022}. For example, in the sentence ``The mayor’s policies led to increased housing costs'', the causal relationship linking the cause (the mayor’s policies) to the effect (increased housing costs) is conveyed by the verb ``led''. In linguistics, causal language is defined as a statement that contains two or more events that are causally linked \citep{solstad_2017}. It can be expressed in many ways, most commonly relying on causal connectives like ``because'', ``therefore'' or ``since'', and through the use of causal verbs like ``cause'', ``affect'' or ``lead''. However, expressions of causality are highly diverse, and they often require inference from context \citep{solstad_2017, Neeleman2012}.

\begin{figure}[h!]
    \centering
    \includegraphics[width=0.8\linewidth]{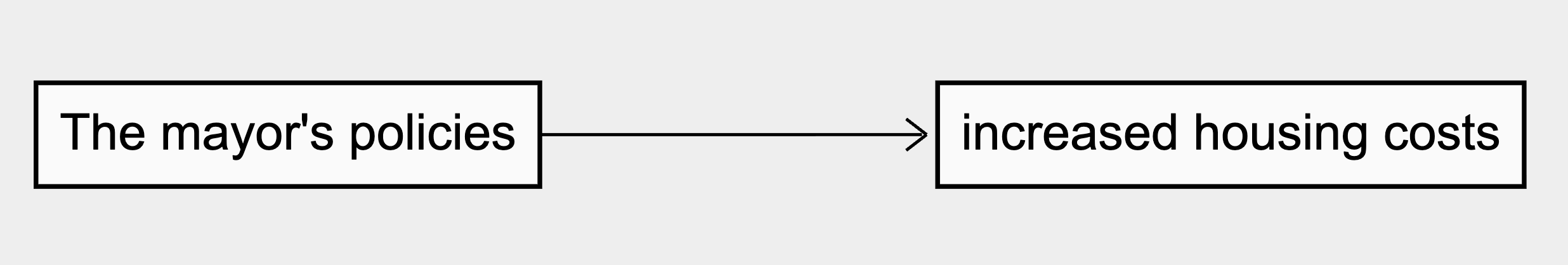}
    \caption{From the sentence ``The mayor's policies led to increased housing costs'' we can identify the cause and effect and represent it in graph form.}
    \label{fig:DAG1}
\end{figure}

Causal language can be systematically identified and analyzed using Natural Language Processing (NLP) tools. By broadly asking the questions ``\textit{What is the stated cause?}'' and ``\textit{What is the stated effect?}'', each causal claim is broken into cause-effect pairs which include a cause, an effect, and the \textit{causal} relationship connecting the events as it's subparts. It is important to highlight that just because a cause and an effect are extracted from a sentence, it does not mean that the causal relationship holds true in the real world. ``Autism is caused by vaccines'', is a causal claim that is known to be scientifically false. However, as a stated causal claim, we can easily identify the cause-effect pair that it is constructed with.

This distinction is important because the variations in claims are what shape how citizens understand, debate and form opinions around the same issues. Political actors often make causal claims that serve strategic purposes like mobilizing support, deflecting blame, or reinforcing group identity, and the claims might be misleading, over-simplifications or false attributions. Therefore, the focus is not on whether claims are true, but on \textit{what} is being claimed as causal, and how claims are constructed across public discourse.

\begin{figure}[h!]
    \centering
    \includegraphics[width=0.5\linewidth]{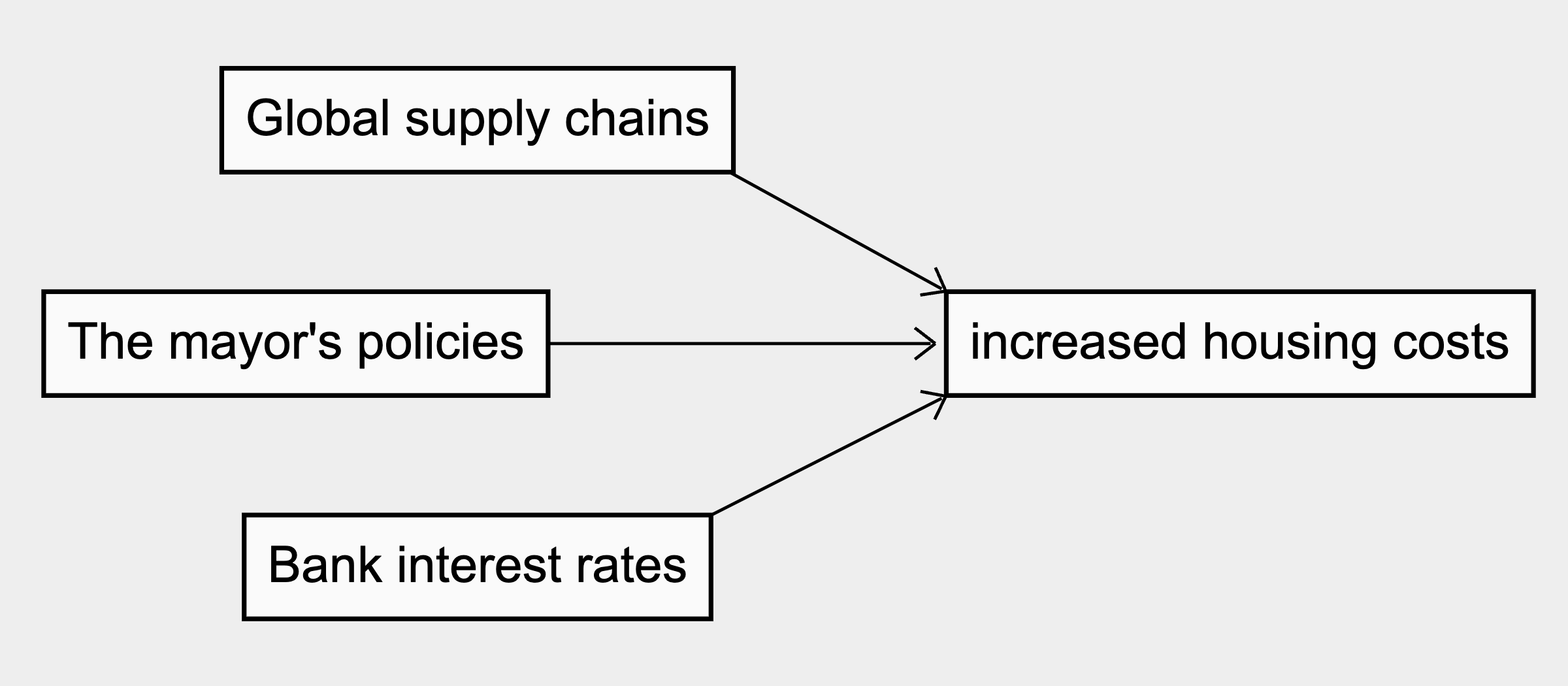}
    \caption{Competing actors frequently offer alternative causal accounts of the same event. Using causal extraction, we can parse a corpus, and extract the causal attributions made around specific issues of interest.}
    \label{fig:DAG2}
\end{figure}

\subsection{Extracting cause-effect pairs from text}
\label{ssec:extracting_cause-effect-pairs-from_text}

To extract structured causal attribution data from unstructured text, I present a framework that relies on fine-tuning Causal Language Models (CLMs) for political text. CLMs are usually pre-trained transformer-based models fine-tuned to capture causal language constructions for classification and span detection. Their outputs consist of (1) a predicted label for the presence of a causal claim, (2) a predicted span for the cause and (3) a predicted span for the effect \citep{dunietz-etal-2017-corpus, drury_survey_2022, tan_unicausal_2023}.

\subsubsection{Causal language annotation}
\label{sssec:causal-language-annotation}

As with other transformer-based models, CLMs need domain-specific annotated data for fine-tuning. I use an annotation codebook adapted from PolitiCAUSE \citep{garcia-corral-etal-2024-politicause}. The codebook follows a semantic annotation framework that focuses on the meaning rather than syntactic or grammatical analysis to label causal claims. Annotation is limited to single sentences or their subparts. Annotators are instructed to assess each sentence and determine whether a sentence contains an explicit causal claim. A sentence is considered causal only if both the cause and its corresponding effect are explicitly expressed within the sentence. For sentences classified as causal, annotators then need to identify and annotate the text spans representing the cause and the effect.

\begin{figure}[h!]
    \centering
    \includegraphics[width=0.7\linewidth]{sentence-examples.png}
    \caption{The first sentence is labeled as ``causal'' because the cause and the effect are explicitly expressed in the sentence. The second sentence is labeled as ``not causal'', because there is no explicit cause expressed within the sentence.}
    \label{fig:example-annotation}
\end{figure}

Annotators are also instructed to include causal verbs in the effect span when the verb's meaning conveys information about the effect. For example, in the sentence ``Recent policy implementations \textbf{increased} housing costs'', the effect is encoded within the verb ``increased'', which implicitly means ``caused the cost to be higher''. Therefore, the verb should be included as part of the effect span during annotation.

\subsubsection{Fine-tuning language models for causal claim extraction}
\label{sssec:Fine-tuning-language-models}

For this framework, I use the first two stages of causal language modeling: (1) sequence classification, and (2) span detection \citep{tan_unicausal_2023}. Most transformer-based language model can be fine-tuned for the task\footnote{Typical examples include \textsc{BERT} and its variations such as \textsc{RoBERTa}, \textsc{DistillBERT}.}. I use \textsc{BERT} \citep{devlin_bert_2019}, as it is publicly available, performs well on causality-related NLP benchmarks, and captures broader contextual information. The annotated sentence are used as the input sequence for fine-tuning. I also employ the Optuna library \citep{optuna_2019} to optimize hyperparameter selection.

\paragraph{Sequence classification} To fine-tune a pretrained BERT model for causal sequence classification, each input sequence $Seq$ is tokenized. A special [CLS] token is prepended and a [SEP] token is appended. During the forward pass the model’s embedding layer maps every token to a learnable embedding vector $\mathbf E_{token_i}$.

The full sequence of embeddings is fed through the BERT encoder, whose weights are fine‑tuned during training. For classification, it takes the final‑layer hidden state at the $\mathbf R_{[CLS]}$ position as a fixed‑length representation of the whole sequence. The embedding $\mathbf R_{[CLS]}$ is passed to a classifier head $f(\cdot)$ that outputs logits ${\hat{y}_{seq}}=f(\mathbf R_{[CLS]})$ for the binary labels causal (1) and non‑causal (0). The logits are converted to class probabilities and the parameters are updated by minimizing negative log-likelihood with respect to the gold labels. Gradients are back‑propagated through both the classifier head and the BERT encoder, completing fine‑tuning.

\begin{figure}[h!]
    \centering
    \includegraphics[width=0.9\linewidth]{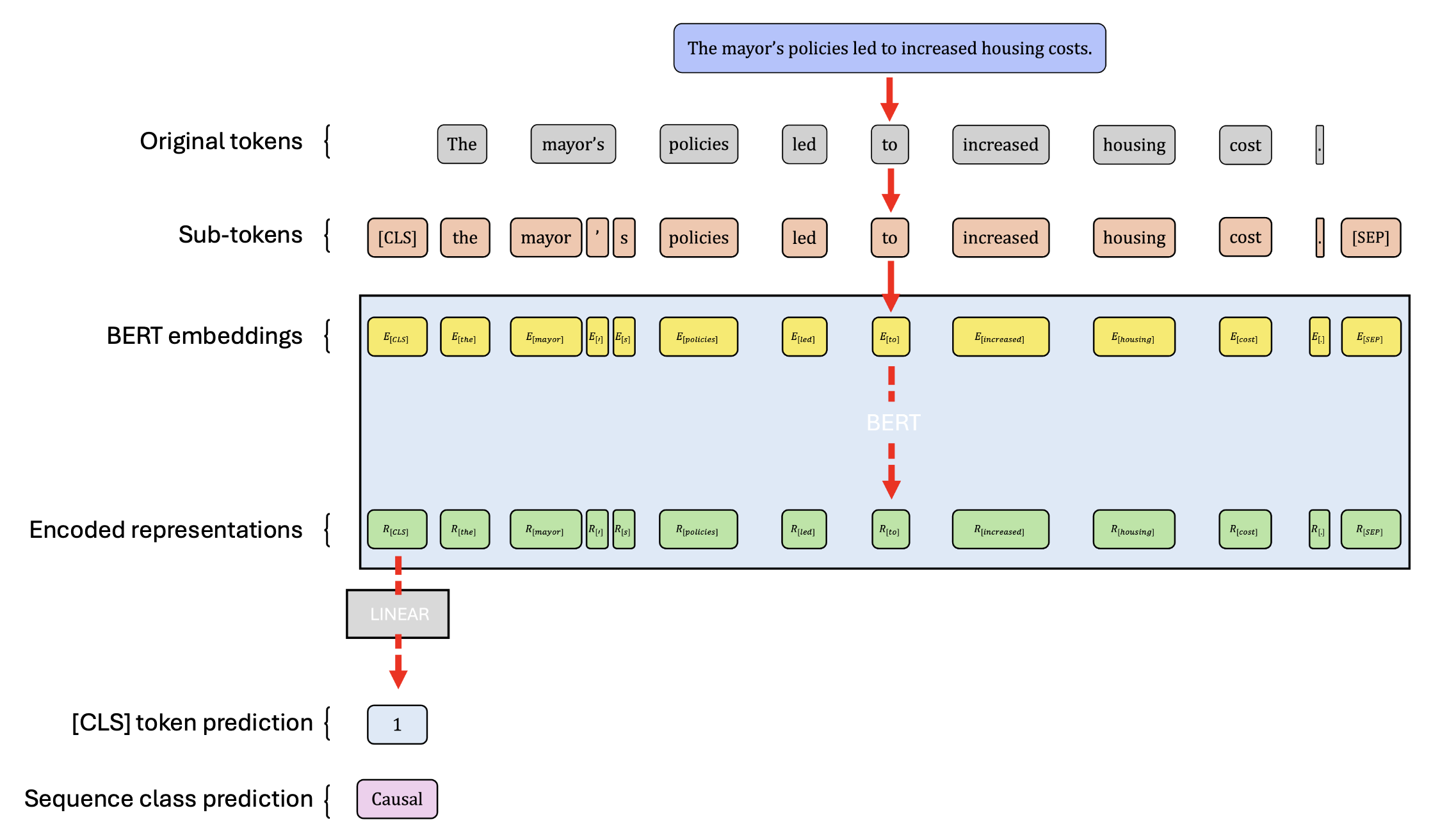}
    \caption{To fine-tune a sequence classification model, a sequence is tokenized into subtokens and the [CLS] and [SEP] tokens are added. The embeddings are pushed through a hidden state. A classification model determines if the sequence is positive or negative by predicting the logits of the sequence embeddings in the [CLS] token.}
    \label{fig:BERT-clf}
\end{figure}

\paragraph{Span Detection} I formulate cause–effect span detection as token classification with IOB2 labels [B‑C (begin‑cause), I‑C (inside‑cause), B‑E (begin‑effect), I‑E (inside‑effect), or O (outside)]. Each input sequence is tokenized and the [CLS] and [SEP] tokens are added. Next, each subtoken is mapped to an embedding vector $\mathbf E_{subtoken_i}$, and all embeddings are passed to the encoder, yielding contextualized hidden states $\mathbf R_{token_i}$. A subtoken‑level classifier head $f(\cdot)$ projects each  $\mathbf R_{subtoken_i}$ to logits $\hat{\mathbf y}_{subtoken_i} = f(\mathbf R_{subtoken_i})$. It then models class membership with a multinomial logistic regression and estimates the parameters minimizing negative log-likelihood against gold labels. To complete the fine‑tuning, the gradients are back‑propagated through both the classifier head and the encoder.

\begin{figure}[H]
    \centering
    \includegraphics[width=0.9\linewidth]{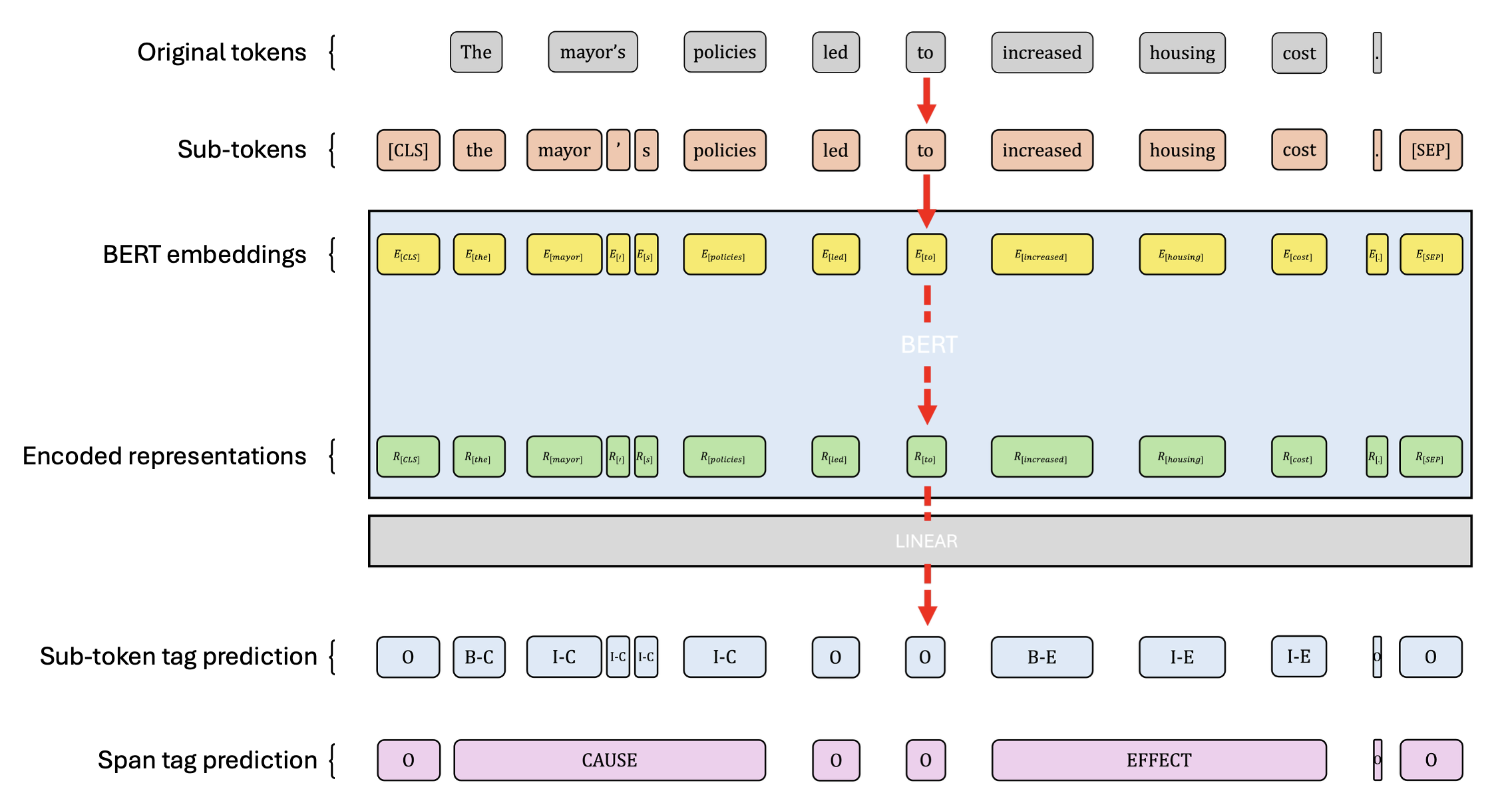}
    \caption{To fine-tune a span detection model a sequence is tokenized and the [CLS] and [SEP] tokens are added. A BERT token classification model assigns a label to each subtoken by feeding their embeddings into a classifier, selecting the label with the highest probability. Spans are then extracted using the IOB2 labeling scheme, where contiguous tokens labeled with 'B-' (begin) and 'I-' (inside) tags form identified spans.}
    \label{fig:BERT-Span}
\end{figure}

\paragraph{Evaluation} I use an out of sample validation set for evaluation. I rely on standard metrics to compute the models performance: Accuracy, Precision, Recall, and F1-score. For classification predictions, these metrics are computed at the sequence level by directly comparing the predicted labels to the corresponding gold labels. For span predictions, a predicted span is considered correct only if its boundaries and entity type exactly match those of a gold span. For each evaluation setting, results are reported using both the macro average, which assigns equal weight to all labels regardless of class frequency, and the weighted average. I use both averages to better assess model performance across labels.

\paragraph{Span reconstruction} To obtain meaningful representations of politically relevant objects, such as actors, events, or actions, I reconstruct each predicted span from its constituent subtokens into complete word-level tokens, and assign a consistent probability estimate for downstream analysis. Directly using the model’s predictions at the subtoken level is impractical as many tokens are split into multiple subtokens. In such cases, a tokens’s probability could be derived from only one of its subtokens or from a pooled value across its constituent subtokens, leading to inconsistencies.

To address this, I implement a two-step procedure. First, I compute probabilities at the span level. Contiguous spans are identified from the predicted IOB2 tags applied to the subtokens. For each span
\begin{equation}
    S = {\text{subtoken}_1, \text{subtoken}_2, \dots, \text{subtoken}_n}
\end{equation}

the averaged span probability is defined as

\begin{equation}
    {P(\hat{S})} = \frac{1}{n}\sum_{i=1}^{n}{\hat{y}_i}
\end{equation}

where $\hat{y}_i$ is the probability assigned to the predicted IOB2 tag of $\text{subtoken}_i$. This yields a single probability representing the model’s confidence in the span as a whole, rather than in individual subtokens.

Second, I assign back the probabilities at the token (word) level. Subtokens are mapped back to their original word-level tokens by doing light text normalization (fixing padding, correcting spacing, restoring hyphens and apostrophes, etc.). The resulting spans now have interpretable word-level tokens, and the span’s average probability 
${P(\hat{S})}$ is assigned uniformly to all tokens within it. This process produces a dataset that uses the span-level probability to assign token-level (per word) probabilities, enabling consistent and interpretable downstream analysis.

\begin{figure}
    \centering
    \includegraphics[width=0.90\linewidth]{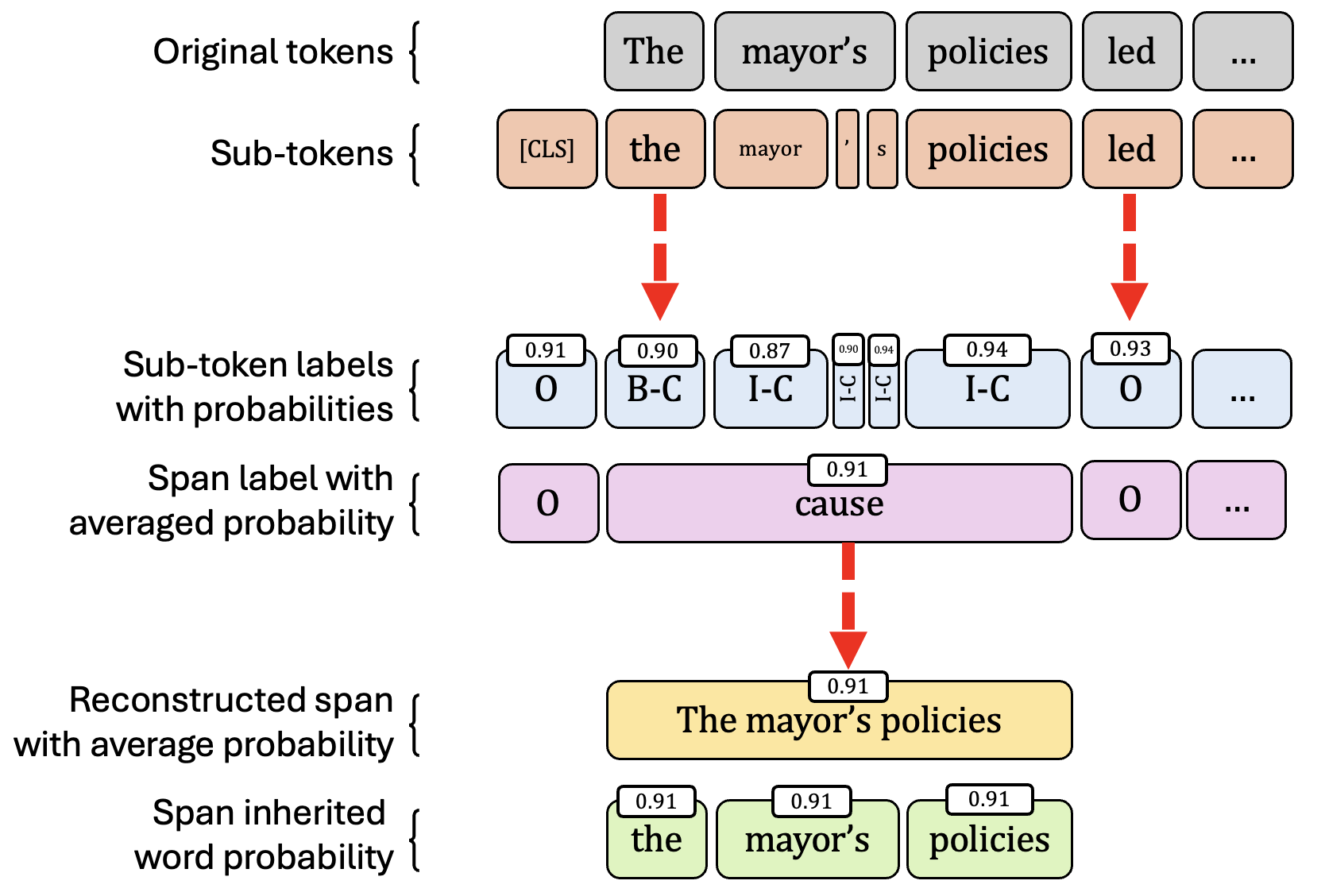}
    \caption{Caption}
    \label{fig:placeholder}
\end{figure}

\section{Examining causal frames in armed conflict reporting}
\label{sec:case-study}

To demonstrate an implementation of the framework, I analyze news headlines from the Israel-Palestine war (ME) and Russia-Ukraine war (EE). I use the CLM framework to examine how media outlets causally frame relevant political actors  when covering armed conflict. For this analysis, causal frames are operationalized as cause-effect pairs. The research question is ``How are political actors causally framed in the media?''.

To study this I select three global news organizations: Al Jazeera (AJ), the BBC, and CNN, and scrape their online websites that cover the conflicts between May 2023 and February 2024. After scarping, I filter out any headlines that do not mention words related to the conflict, based on a hand-crafted dictionary. This yields a final corpus of 4,993 headlines (See Appendix \ref{ssec:appendix-corpus} for full statistics). 

\subsection{Cause-Effect pair extraction}
\label{ssec:cause-effect-pais-extraction}

To obtain all the cause-effect pairs in the corpus, I begin by annotating a randomly selected subset of 1,238 headlines weighted across conflict and media outlet. For this, 8 research assistants were hired and trained to annotate, each sentence was annotated three times. I aggregated the causal label by majority voting of each sentence, and the causal spans using Overlap-Based Consensus\footnote{To quantify the agreement between spans from different annotators, I used Intersection over Union and a threshold of $\tau$ = 0.5.}. With the annotation process completed, I divide the annotated data into train, test and validation sets (70\% - 15\% - 15\%), and use the training and testing sets to complete fine-tuning for each CLM (See Appendix \ref{ssec:appendix-ttv-splits} and Appendix \ref{sec:appendix-ft-strategy} for training data, model selection and fine-tuning details).

The Sequence Classification model demonstrated strong performance in identifying causal relationships (Table \ref{tab:results_Clf_BERT}). Overall, the model has high macro and weighted F1-scores of 81\% and 83\%, respectively, and an overall accuracy of 83\%, which indicates a good model for sequence classification.

\begin{table}[h]
\captionsetup{font=scriptsize}
\begin{center}
    \begin{tabular}{lrrrr}
\toprule
 & precision & recall & accuracy & f1-score \\
\midrule
CAUSAL       & 0.891 & 0.855 &       & 0.872 \\
NOT CAUSAL   & 0.727 & 0.787 &       & 0.756 \\
macro avg    & 0.809 & 0.821 & 0.832 & 0.814 \\
weighted avg & 0.837 & 0.832 & 0.832 & 0.834 \\
\bottomrule
\end{tabular}
    \caption{Results of Sequence Classification BERT on validation data}
    \label{tab:results_Clf_BERT}
    \end{center}
\end{table}

Furthermore, the Span Detection model shows excellent performance (Table \ref{tab:results_NER_BERT}). It achieves precision and recall scores above 90\% for both span types, with F1-scores of 92\% for Cause and 94\% for Effect. Averaged metrics consistently yield F1-scores of 93\%, indicating highly reliable span detection across categories. 

\begin{table}[h]
\captionsetup{font=scriptsize}
\begin{center}
    \begin{tabular}{lrrrr}
\toprule
 & precision & recall & f1-score \\
\midrule
CAUSE        & 0.90 & 0.95 & 0.92 \\
EFFECT       & 0.93 & 0.95 & 0.94 \\
micro avg    & 0.91 & 0.95 & 0.93 \\
macro avg    & 0.91 & 0.95 & 0.93 \\
weighted avg & 0.92 & 0.95 & 0.93 \\
\bottomrule
\end{tabular}
    \caption{Results of Span detection BERT on validation data}
    \label{tab:results_NER_BERT}
    \end{center}
\end{table}

\subsection{Validation}

\begin{figure}[!h]
     \centering
     \includegraphics[width=0.8\linewidth]{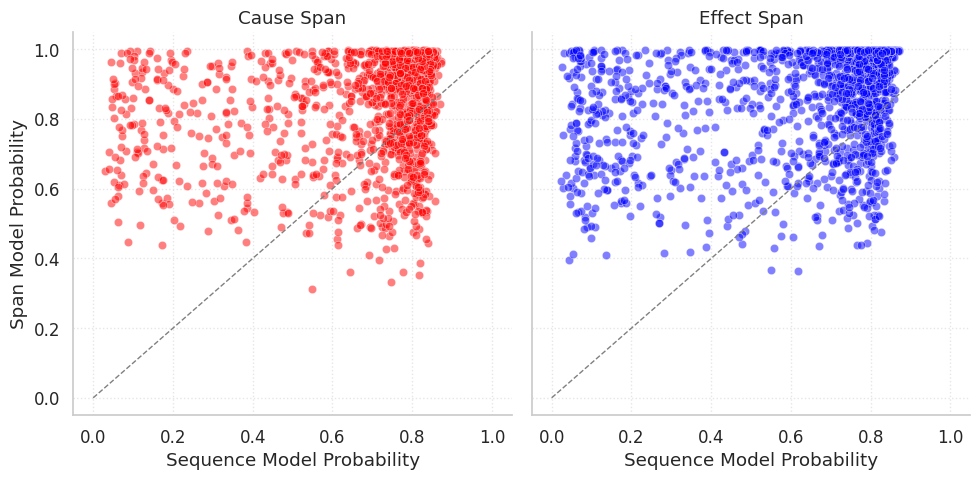}
     \caption{The plots show the cause span probability and the effect span probability plotted against the sequence probability. The show a fairly equal performance on detecting causal language constructions.}
     \label{fig:scatter-plot-probs}
\end{figure}

I investigate the models' validity by comparing if both models are capturing the same construct. In Figure \ref{fig:scatter-plot-probs}, the scatter plots compare the sequence classification model probabilities against span detection model probabilities, according to each span, by sequence. The plot indicates a high degree of agreement between the two models, particularly when both assign high probabilities. This suggests that when the sequence model confidently predicts the presence of a causal relation, the span model identifies the corresponding cause and effect spans with similarly high confidence. There are no big deviations in the mid-probability range, suggesting that the span detection model has a fairly equal performance for both the cause and effect section of each sentence. We also see that there is more confidence in the span than in the sequence model, where the highest probability reaches maximum levels at around 0.80, while the span model is highly confident, with values touching the limit of 1.0.

The overall result metrics and validation analysis indicate that the models have good performance, and that they encode similar causal semantics (Further validation is available in Appendix \ref{ssec:appendix-validation-analysis}). I deploy the CLMs to annotate the rest of the corpus. Then, I reconstruct the spans and obtain the token(word)-level estimated probabilities for being in the cause or effect.

\subsection{How are political actors causally framed?}

To identify how political actors are framed according to conflict and media source, I measure actor‐level associations with cause versus effect attribution by extending the log‐odds framework of \citet{Monroe_Colaresi_Quinn_2017} to incorporate the probabilistic outputs of the span detection model (Appendix \ref{sec:appendix_tokens_stats_breakdown}). To do this, I sum the token-level probabilities of each token to form soft counts for cause and effect, by media source and conflict (group); I then compute class totals over all tokens, estimate each token’s total share across spans, and apply an informative Dirichlet prior that allocates pseudo‐counts to ``present'' and ``absent'' cells based on that share. From the resulting smoothed $2 \times 2$ table, I compute odds for each span, odds ratio for the cause vs. effect prevalence ($\delta$), and the log‐odds ratio (log($\delta$)). As well as derive the Wald standard errors (se), confidence intervals (CIs), and p-values on the log scale. 

\begin{figure}[h!]
    \centering
    \includegraphics[width=1\linewidth]{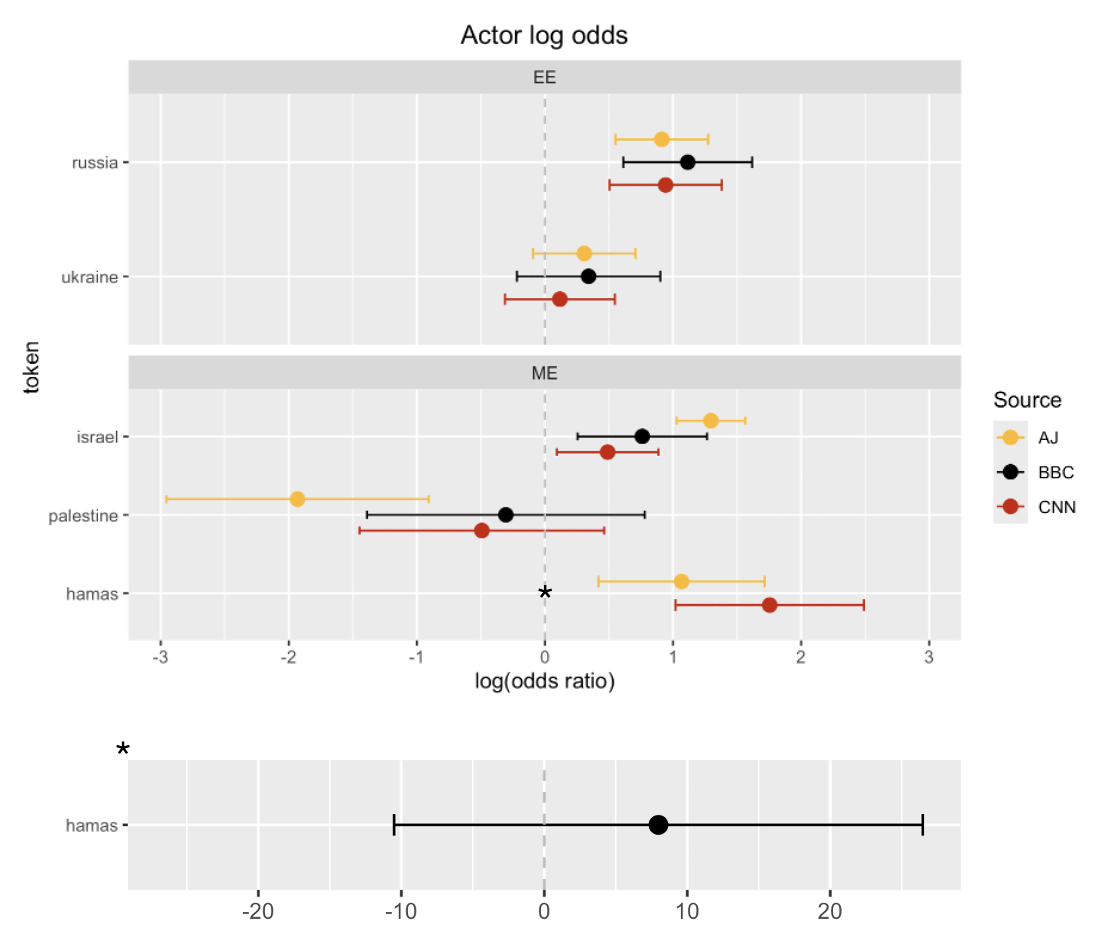}
    \caption{Each actor's log odds ratio are plotted by source. We observe that AJ causally frames Palestine as systemically being in effects to a greater degree than any other actor. I add the BBC's hamas value below as it has a different proportion as rare-token artifact. For a full comparison see Appendix.}
    \label{fig:log-odds-actors}
\end{figure}

I obtain the log odds for five salient actors in the conflicts (Russia, Ukraine, Hamas, Israel and Palestine)\footnote{I aggregate actor tokens into singular groups for ease of interpretation. For example, ``russia'', ``russian'', ``russians'' are aggregated as ``russia''.}. Figure \ref{fig:log-odds-actors} plots token-level log-odds (cause vs. effect) with 95\% CIs. Comparing the two conflicts shows differences in both central tendency and dispersion. In EE, Russia exhibits a uniformly cause-sided framing across outlets ($0.50 < \text{log}(\delta) < 1.11$; all $ p < 0.05$), whereas Ukraine does not ($-0.31 < \text{log}(\delta) < 0.90$; none significant (n.s.)). This suggests a stable, cross-outlet pattern in which Russia is consistently framed as a causal agent.

In ME, framing varies by outlet. Hamas is generally cause-sided ($1.07 < \text{log}(\delta) < 7.99$, significant for AJ and CNN $p < 0.05$ ), but BBC shows no effect-span mentions, producing large estimates with wide, non-significant intervals (a known artifact when one span has zero counts.) Palestine is less consistent: only AJ shows effect-sided framing ($-2.96< \text{log}(\delta) < -0.91$; $p < 0.05$), while BBC and CNN straddle zero ($-1.45 < \text{log}(\delta) < 0.78$; n.s.). Israel is consistently cause-sided across outlets ($0.09 < \text{log}(\delta) < 1.56 $, all $p < 0.05$) Overall, EE exhibits a coherent ``Russia-as-cause'' narrative with relatively tight, significant intervals, while ME displays outlet-specific patterns, especially for Hamas and Palestine which reflects more heterogeneous framing (See full actors statistics in Appendix \ref{ssec:full_corpus_actor_analysis}.)

\newpage

\subsubsection{\textit{Who} causes \textit{what}?}

To deepen the analysis of causal frames, I look at how causal claims are constructed across the corpus to answer the questions: ``What are the main effects attributed to political actors?'' and  ``What are the main causes affecting political actors?''. To do this, I simply subset the dataset according to the causal claim I want to evaluate. For example, to examine how the media is causally framing attacks, I select the cause-effect pairs that contain the token ``attack'' across both span, and then calculate log odds for the actors I want to compare. By doing this, I effectively ``hold constant'' the desired token of analysis in the text of interest, and measure the actor cause-effect pair variations in the subset of the corpus.

\begin{figure}[h!]
\centering
\includegraphics[width=0.9\linewidth]{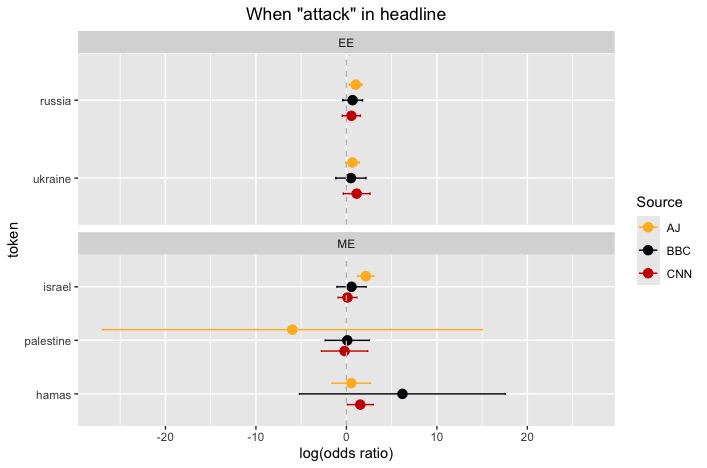}
    \caption{The word attack appears in the headline}
    \label{fig:attack-in-text}
\end{figure}

In Figure \ref{fig:attack-in-text}, we observe the estimated log-odds after subsetting to only analyze events that refer to an attack. This further provides evidence of framing attribution by estimating token-level log-odds for ``who attacks?'' (cause) and ``who is attacked?'' (effect). In EE, the only statistically significant result is AJ’s framing of Russia as a causal agent ($0.33 < \text{log}(\delta)< 1.73$; $p<0.05$); all other actor–outlet estimates are non-significant. In ME, patterns are more differentiated: CNN frames Hamas as an attacker ($0.09 < \text{log}(\delta)< 2.98$; $p<0.05$); AJ frames Israel as an attacker ($1.25 < \text{log}(\delta)< 3.04$; $p<0.05$); and AJ’s portrayal of Palestine is effect-sided but highly uncertain ($-26.96 < \text{log}(\delta)< 15.01$; n.s.), reflecting sparse cause-span mentions. Similarly, the BBC has no effect-span mentions of Hamas, yielding large, imprecise estimates ($-5.19 < \text{log}(\delta)< 17.58$; n.s.). In short, EE shows limited, AJ-driven cause-framing of Russia, whereas ME exhibits outlet-specific attacker/victim framing: CNN (and, in point estimates, BBC) cast Hamas as attacker; AJ casts Israel as attacker and Palestine as attacked (See Appendix \ref{ssec:appendix_attack_in_headline} for full statistical analysis).

This analysis shows the relevance of using the CLM framework to study how media causally frames conflicts. The description of a relevant actors in the cause side or effect side of headlines directly informs the agency and responsibility of different agents across conflicts, providing more detailed estimates of causal attribution in media. This results are pertinent for the study of ``responsibility attribution'' and ``character frames'', both causal elements that have a direct implications on how media shapes news through the use (or lack of) causal attribution for certain political actors.

\section{Limitations}
This framework has certain constraints to consider. First, causal language is inherently contextual and often difficult to annotate reliably. High-quality annotations are essential for training effective models, yet establishing consistent annotation standards requires specialized training. Additionally, the semantics of causal expressions are domain-specific: a model trained on headlines may not generalize well to other text genres.

As for the results, while they offer a useful demonstration of the proposed method, they should be interpreted with caution. First, the analysis does not account for temporal variation: the corpus spans from May 2023 to February 2024, a period during which systematic differences across conflicts in the media coverage are likely to emerge due to different stages of each conflict. Second, topic frequency is not controlled for, meaning that comparisons across events are based on absolute frequencies rather than normalized measures. This may influence the apparent salience of causal language in coverage of particular events.

\section{Discussion}
\label{sec:discussion}

\subsection{Semantics as an analytical tool}
\label{ssec:semantics}
The CLM framework introduces semantics, and labels text based on a linguistic relationship that is known, erasing the interpretative step of that is usually needed to connects different data points. Thus, this method reduces polysemy and eliminates an often necessary interpretative step when classifying data by using the linguistic representation of a unique semantic relationship. The advantage has been leveraged in other recent text-as-data methodologies such as \citet{Ash2024}, showing how linguistic sensitivity adds depth to the interpretation of textual data, without sacrificing scale.

\subsection{Annotation rules}
\label{ssec:annotation-rules}
The choice of annotation codebook is a key methodological decision especially since this method relies fully on the annotation of a linguistic relationship. Currently, different annotation approaches exist \citep{dunietz-etal-2017-corpus, mirza_event_2021, garcia-corral-etal-2024-politicause}, and they vary in the extent to which they prioritize syntactic rigor versus interpretive flexibility. Since my primary interest lies on rhetorical interpretation, instead of formal linguistic constructions, I choose a semantic framework. However, more constrained annotation schemes are useful in other contexts, particularly those aiming to distinguish between different types of causal language. For example, when distinguishing between blame or claim statements, or when analyzing the use of causal or correlational language when studying scientific evidence in policy documents.

\subsection{Language model selection}
\label{ssec:LM-selection}
In this paper I use \textsc{BERT}, which has demonstrated strong performance in similar tasks for Computational Linguistic use cases \citep{tan_unicausal_2023, drury_survey_2022}. Nonetheless, models choice should be done according to specific needs, such as language considerations or domain particularities, which can yield higher performance for the specific use cases. Furthermore, recent advances in generative AI, such as the introduction of the GPT model families or Llama, provide a promising avenue for scalable and cost-effective annotation of causal language. Preliminary research suggests that these large language models are capable of producing high-quality outputs in causal relation extraction tasks \citep{takayanagi2024chatgptfuturecausaltext, hobbhahn2022investigating, shukla_et_al_2023, gao-etal-2023-chatgpt}.

\section{Conclusion}
\label{sec:conclusion}
The framework presented here provides a systematic way to detect and analyze causal claims, enabling new kinds of questions and comparisons for the study of causal attributions and explanations in political discourse. By offering a generalizable, low-cost tool for extracting causal claims from text, this framework unlocks a rich but underutilized source of data in political science. After modeling a corpus for causal language, the pipeline relies on nothing more complex than log-odds ratios and Dirichlet smoothing; the power comes from rooting these established statistical tools in robust linguistic insights to bring the analysis to life. Causal explanations, long treated as anecdotal or ad hoc, can be measured and analyzed with rigor, bringing us closer to understanding how political meaning is constructed and contested in the public sphere.

\bibliography{costum}

\newpage

\appendix

\section{Data}
\subsection{Corpus}
\label{ssec:appendix-corpus}

For the case study, I constructed a corpus by scraping the online news pages for AJ, BBC, and CNN. I gathered all of the data in the relevant section (Europe, Middle East, or Conflict specific sections), did posthoc filtering to drop unrelated content using a dictionary for each conflict: (gaza OR palestine OR israel OR hamas OR west bank) and (ukraine OR russia OR kiev OR kyiv OR putin OR zelensky OR moscow).

\begin{table}[h!]
\captionsetup{font=scriptsize}
\centering
\begin{tabular}{@{}llr@{}}
\toprule
Region & Source & Count (Proportion)\\
\midrule
ME & AJ & 1,251 (0.48) \\
& BBC & 792 (0.30) \\
& CNN & 567 (0.22) \\
\midrule
EE& AJ & 1,018 (0.43) \\
& BBC & 784 (0.33) \\
& CNN & 581 (0.24) \\
\bottomrule
\end{tabular}
\caption{Statistics for the corpus of all collected news articles headlines. To specify conflict, I use the ``Region'' variable, where ME stands for the Israel-Palestine war in the Middle East and EE for the Russian-Ukraine war in Eastern Europe. ``Source'' refers to the news organization (AJ, BBC or CNN). The values to the far right are the total number of headlines with the proportion relative to region in parenthesis.}
\label{tab:Full-Corpus-Statistics}
\end{table}

\subsection{Annotation example}
\label{ssec:appendix-annotation-example}

I hired 8 research assistants (RA) for the data annotation, all RAs were graduate students from social science programs. They were trained and asked to complete a test set before labeling the final dataset. Each sentence was annotated three times to gather majority voting for the labels. To annotate the text, I used a combination of Potato annotation software \citep{pei2022potato}, AWS to host the web app, and Prolific for annotator management.

\begin{figure}[h!]
    \centering
    \includegraphics[width=0.5\linewidth]{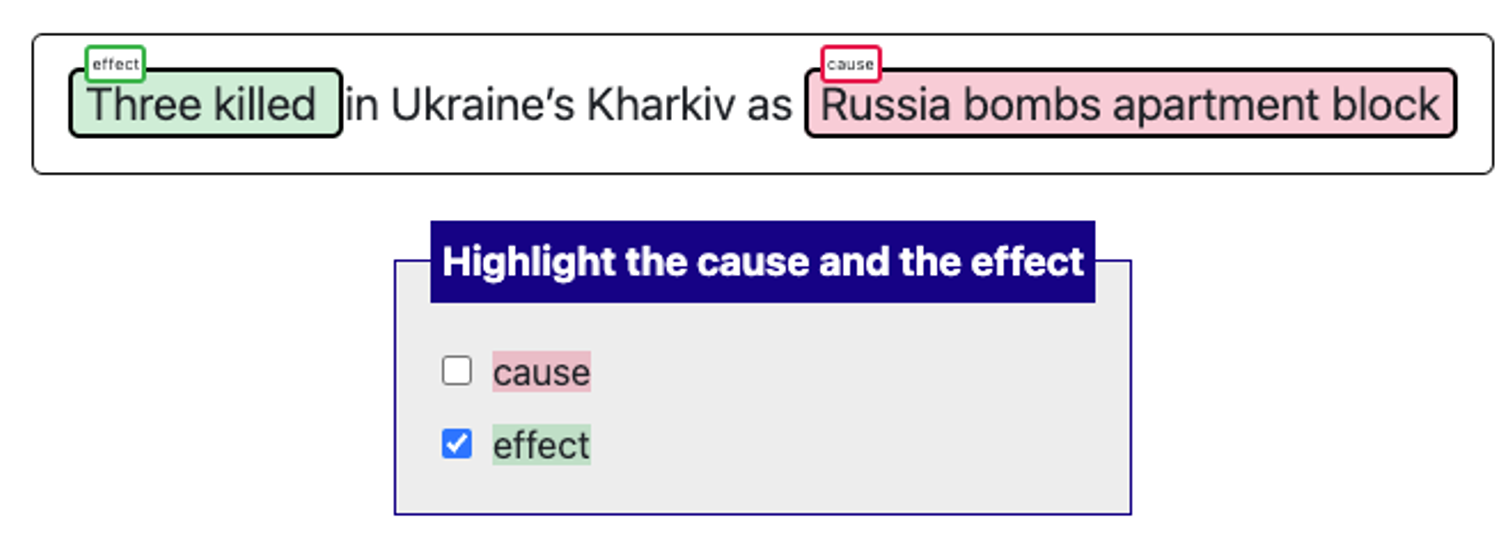}
    \caption{Annotation of a headline using Potato}
    \label{fig:annotation-example}
\end{figure}

\hfill

\subsection{Train Test Validation splits}
\label{ssec:appendix-ttv-splits}

To train the models I split into three splits (75-15-15) and gathered into two subsets saved as Hugging Face datasets as \href{https://huggingface.co/datasets/pgarco/CausalPalUkr_Train_Test}{pgarco/CausalPalUkr\_Train\_Test} and \href{https://huggingface.co/datasets/pgarco/CausalPalUkr_Val}{pgarco/CausalPalUkr\_Val}. The Train-Test subset was used for training the models calculating loss during training. The validation subset was used to obtain out-of-sample performance metrics for the final fine-tuned models (Reported in section \ref{ssec:cause-effect-pais-extraction}. The Table \ref{tab:Annotated-set-Statistics} reports the statistics of both subsets. 

\begin{table}[h!]
\captionsetup{font=scriptsize}
\centering
\begin{tabular}{@{}lr@{}}
\toprule
Set &  Count (Proportion)\\
\midrule
Train &  862 (0.70) \\
Test  &  191 (0.15) \\
Val   &  185 (0.15) \\
Total & 1238 (1.00) \\
\bottomrule
\end{tabular}
\caption{Statistics for the human annotation subset of the corpus. A split of 70\% - 15\% - 15\%, was selected where the covariates are weighted evenly across each set.}
\label{tab:Annotated-set-Statistics}
\end{table}

\newpage

\section{CLM Fine-tuning strategy}
\label{sec:appendix-ft-strategy}

I finetune two CLMs: (1) sequence classification and (2)  span detection. I selected \textsc{BERT}, a Bi-directional Encoder Representations from Transformers model \citep{devlin_bert_2019}, which has an ``encoder-only'' transformer architecture, with multiple self-attention heads, trained via Masked Language Modeling and Next Sentence Prediction tasks. I followed the standard procedure for NLP classification and span detection tasks and used Hugging face’s base tokenizer and configuration for each model\footnote{\url{https://github.com/huggingface/transformers/tree/v4.32.1/src/transformers/models/bert}}. Furthermore, I used a Cross Validation K-Fold strategy to analyze the models' performances across different splits, and Optuna for hyperparameter search.

\subsection{Text Normalization, pre-processing and cleaning after predictions}
\label{ssec:appendix_text_normalization_preprocessing}
After using the fine-tuned CLMs to machine-annotate the remaining corpus the span prediction tokens are converted back into string representations for each cause and effect span. The string representations were preprocessed by reattaching split tokens, hyphenate proper names, and consolidated synonyms into one single token. Arab prefixes such as ``Al'' were merged into a single concatenated token from ``al Shifa'' to ``alshifa''. Everything was turned to lowercase. 

I conducted a manual inspection on the resulting token list, and created a correction dictionary develop to fix tokens that are relevant for the analysis according to domain importance. For example, unifying ``child'',``children'', to ``children'' for ease of analysis. This process was done abductively until a solid pipeline for data cleaning and preprocessing was configured for the whole corpus. The dictionaries for text processing are in the text-preprocessing files.

\newpage

\subsubsection{Sequence model information and training details}
\label{ssec:Seq_model_info_train}

\begin{table}[h!]
\centering
\resizebox{\textwidth}{!}{\begin{tabular}{lll}
Category & Parameter & Value \\
\toprule
 \textbf{Model Architecture}& Base model & bert-base-uncased\\
 & Architecture & BertForSequenceClassification \\
 & Number of labels & 2 \\
 & Hidden size & 768 \\
 & Intermediate size & 3072 \\
 & Hidden layers & 12 \\
 & Attention heads & 12 \\
 & Activation & GELU \\
 & Dropout & 0.1 \\
 & Tokenizer & BertTokenizer (max length=512,\\
  & & padding/truncation) \\
\midrule
 \textbf{Dataset}& Source & CausalPalUkr\_Train\_Test \\
 & Task type & Single-label classification \\
 & Train/test split & 75-15 \\
 & Preprocessing & Lowercase, truncate/pad to 512,\\
 & & [CLS]/[SEP] tokens \\
\midrule
\textbf{Optimization} & Optimizer & AdamW (adamw\_torch) \\
 & Learning rate & 3e-5 \\
 & Scheduler & Linear decay, 0 warmup steps \\
 & Weight decay & 0.0 \\
 & Betas & (0.9, 0.999) \\
 & Epsilon & 1e-8 \\
 & Batch size / device & 16 \\
 & Gradient accumulation & 1 \\
 & Max grad norm & 1.0 \\
 & Epochs & 3 \\
 & FP16/BF16 & Disabled \\
\midrule
\textbf{Reproducibility} & Seed & 42 \\
 & Data seed & None \\
 & Dataloader drop last & False \\
 & Dataloader workers & 0 \\
 & FP16 full eval & False \\
\midrule 
\textbf{Training Environment} & Transformers & 4.50.3 \\
 & PyTorch & 2.6.0+cu124 \\
 & Datasets & 3.5.0 \\
 & Tokenizers & 0.21.1 \\
 & Hardware & 1x NVIDIA A100 40GB\\
\midrule
\textbf{Evaluation} & Metrics & Accuracy \\
 & Validation freq & after each epoch\\
 \midrule
\textbf{HF model} & \href{https://huggingface.co/pgarco/CausalPalUkr_seq}{pgarco/CausalPalUkr\_seq}\\
\bottomrule
\end{tabular}}
\label{tab:appendix_seq_reproducibility}
\caption{Sequence model information and training details}
\end{table}

\newpage

\subsubsection{Token classification model information and training details}
\label{ssec:Tok_model_info_train}

\begin{table}[h!]
\centering
\resizebox{\textwidth}{!}{\begin{tabular}{lll}
\textbf{Category} & \textbf{Parameter} & \textbf{Value} \\
\toprule
\textbf{Model Architecture} & Base model & bert-base-uncased \\
 & Architecture & BertForTokenClassification \\
 & Hidden size & 768 \\
 & Intermediate size & 3072 \\
 & Hidden layers & 12 \\
 & Attention heads & 12 \\
 & Activation & GELU \\
 & Dropout & 0.1 \\
 & Tokenizer & BertTokenizer (max length=512,\\
 & & padding/truncation) \\
 & Number of labels & 5 \\
 & Label mapping & \{``O'': 0, ``B-C'': 1, ``I-C'': 2,\\
 & & ``B-E'': 3, ``I-E'': 4\} \\
\midrule
\textbf{Dataset} & Name/source & CausalPalUkr\_Train\_Test\\
 & Task type & Token Classification / NER \\
 & Train/val/test split & 75-15\\
 & Preprocessing & Lowercase, truncate/ \\
 & & pad to 512, first subword labeling \\
\midrule
\textbf{Optimization} & Optimizer & AdamW (adamw\_torch) \\
 & Learning rate & 7e-5 \\
 & Scheduler & Linear decay, 36 warmup steps \\
 & Weight decay & 0.01 \\
 & Betas & (0.9, 0.999) \\
 & Epsilon & 1e-8 \\
 & Batch size / device & 16 \\
 & Gradient accumulation & 1 \\
 & Max grad norm & 1.0 \\
 & Epochs & 10 \\
 & FP16/BF16 & Disabled \\
\midrule
\textbf{Reproducibility} & Seed & 42 \\
 & Data seed & None \\
 & Dataloader drop last & False \\
 & Dataloader workers & 0 \\
 & FP16 full eval & False \\
\midrule 
\textbf{Training Environment} & Transformers & 4.50.3 \\
 & PyTorch & 2.6.0+cu124 \\
 & Datasets & 3.5.0 \\
 & Tokenizers & 0.21.1 \\
 & Hardware & 1x NVIDIA A100 40GB\\
\midrule
\textbf{Evaluation} & Metrics & Token-level Precision\\
 & Validation freq & after each epoch \\
  \midrule
\textbf{HF model} & \href{https://huggingface.co/pgarco/CausalPalUkr_token}{pgarco/CausalPalUkr\_token}\\
\bottomrule
\end{tabular}}
\label{tab:appendix_token_reproducibility}
\caption{Token classification model information and training details}
\end{table}

\newpage

\subsection{Validation analysis}
\label{ssec:appendix-validation-analysis}
To further assess the agreement between the models, I use Bland–Altman plots for each span type, to compare the average predicted probabilities of each model against their differences. Overall, I observe that the span model tends to assign higher probabilities than the sequence model, especially for lower-probability spans. This results in a consistent positive bias, indicating again that the models are not directly interchangeable. For cause spans, the average bias is approximately +0.2. The majority of points fall within the 95\% limits of agreement, and very few span predictions fall below zero difference, suggesting the sequence model rarely assigns higher probabilities than the span model. For effect spans, the mean bias is slightly higher (+0.2 to +0.25) with greater vertical spread, indicating more variability and less consistent agreement.

\begin{figure}[!h]
     \centering
     \includegraphics[width=0.8\linewidth]{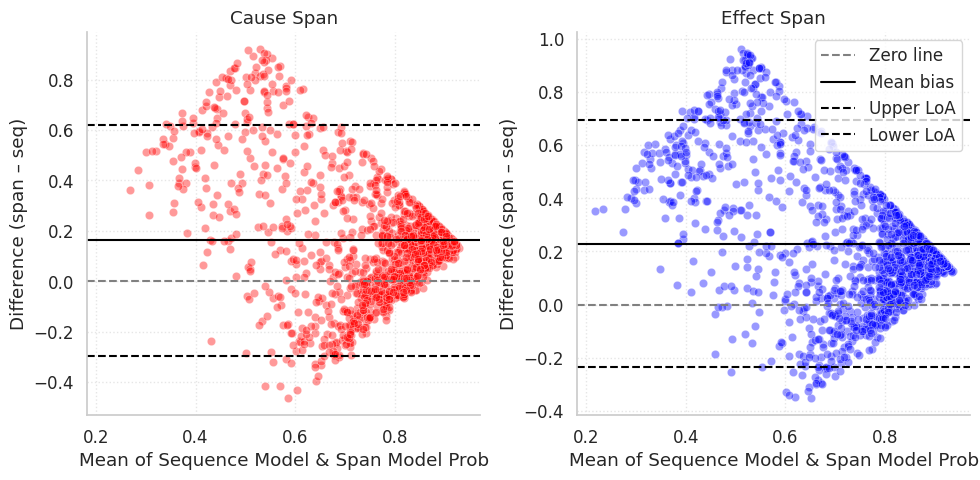}
     \caption{Bland–Altman plots comparing span model and sequence model probabilities for cause and effect spans. The solid line indicates the mean bias (systematic difference), while dashed lines represent the 95\% limits of agreement. For both span types, the span model tends to assign higher probabilities, with slightly more variability in effect spans.}
     \label{fig:bland-altman-plot}
 \end{figure}

The validation analyses indicates that both models are capturing a shared latent signal related to the presence and location of causal relationships in the text, despite differing training approaches. However, the Bland–Altman plots reveal a systematic bias in confidence: the span-based model consistently reports higher confidence than the sequence-level model. This likely reflects differences in information encoding, with the span model having more localized access to linguistic cues and the sequence model adopting a more conservative, aggregated uncertainty approach. 

\newpage

\section{Token-Level Log-Odds with Background-Informed Priors}
\label{sec:appendix_tokens_stats_breakdown}

\paragraph{Setup.}
Every token $w$ is associated with a span class $S\in\{\text{cause},\text{effect}\}$, a probability in $[0,1]$ reflecting confidence in that span assignment, and a group $g$ (e.g., region $\times$ source). Analysis is performed per $(g,w)$.

\paragraph{Soft counts and class totals.}
For each span $S$ and group $g$, define the \emph{soft present count} of token $w$ as the sum of its span probabilities across all of its occurrences in that $(S,g)$:
\[
y_{S,g,w}\;\;=\;\;\text{(sum of span probabilities of token $w$ within span $S$ and group $g$)}
\]
Define the \emph{class total} in $(S,g)$ as the sum of span probabilities across \emph{all} tokens in that span and group:
\[
N_{S,g}\;\;=\;\;\text{(sum of span probabilities of all tokens within span $S$ and group $g$)}
\]

\paragraph{Background share (per group).}
The background prevalence of token $w$ in group $g$ is
\[
p_{\mathrm{background},g,w}\;=\;\frac{y_{\mathrm{cause},g,w}+y_{\mathrm{effect},g,w}}{\,N_{\mathrm{cause},g}+N_{\mathrm{effect},g}\,}\;\in[0,1]
\]

\paragraph{Informative (Monroe) prior.}
Choose a prior strength $\alpha_0>0$ and split it between the ``present'' and ``absent'' cells according to the background share:
\[
\alpha_{\mathrm{present},g,w}=\alpha_0\,p_{\mathrm{background},g,w}
\qquad
\alpha_{\mathrm{absent},g,w}=\alpha_0-\alpha_{\mathrm{present},g,w}
\]

\paragraph{Smoothed $2\times2$ table.}
Using complements to the class totals, construct for $(g,w)$:
\[
\begin{aligned}
a &= y_{\mathrm{cause},g,w} \;+\; \alpha_{\mathrm{present},g,w} &
b &= \bigl(N_{\mathrm{cause},g}-y_{\mathrm{cause},g,w}\bigr) \;+\; \alpha_{\mathrm{absent},g,w}\\[2pt]
c &= y_{\mathrm{effect},g,w} \;+\; \alpha_{\mathrm{present},g,w} &
d &= \bigl(N_{\mathrm{effect},g}-y_{\mathrm{effect},g,w}\bigr) \;+\; \alpha_{\mathrm{absent},g,w}
\end{aligned}
\]

\paragraph{Define span-wise odds and the odds ratio.}
\[
\text{odds}_{\mathrm{cause}}=\frac{a}{b},\qquad
\text{odds}_{\mathrm{effect}}=\frac{c}{d},\qquad
\mathrm{OR}=\frac{\text{odds}_{\mathrm{cause}}}{\text{odds}_{\mathrm{effect}}}
\]
\paragraph{Calculate log-odds}
\[
\delta=\log(\mathrm{OR})
\]

\paragraph{Wald inference on the log scale.}
Use the standard Wald approximation:
\[
\mathrm{SE}(\delta)=\sqrt{\frac{1}{a}+\frac{1}{b}+\frac{1}{c}+\frac{1}{d}},\qquad
z=\frac{\delta}{\mathrm{SE}(\delta)}.
\]
For confidence level $1-\alpha$ with standard normal critical value $z_{1-\alpha/2}$:
\[
\text{CI}_{1-\alpha}(\delta)=\delta \pm z_{1-\alpha/2}\,\mathrm{SE}(\delta),
\qquad
p\text{-value}=2\bigl(1-\Phi(|z|)\bigr),
\]
where $\Phi$ is the standard normal CDF. Exponentiating the limits yields the CI on the odds-ratio scale.

\newpage

\section{Results figures and table}

\subsection{Actor log-odds analysis on full corpus}
\label{ssec:full_corpus_actor_analysis}

\begin{table}[H]
\captionsetup{font=scriptsize}
    \centering
\begin{tabular}{lllrrrr}
\toprule
location & tokens & source & log($\delta$)  & CI lower & CI upper & p-value \\
\midrule
 &  & AJ & 0.91 & 0.55 & 1.27 & 0.00 \\
EE & \textit{russia} & BBC & 1.11 & 0.61 & 1.62 & 0.00 \\
 &  & CNN & 0.94 & 0.50 & 1.38 & 0.00 \\
 &  & AJ & 0.31 & -0.09 & 0.71 & 0.13 \\
EE & \textit{ukraine} & BBC & 0.34 & -0.22 & 0.90 & 0.23 \\
 &  & CNN & 0.12 & -0.31 & 0.55 & 0.59 \\
 & & AJ & 1.07 & 0.42 & 1.72 & 0.00 \\
ME & \textit{hamas} & BBC & 7.99 & -10.51 & 26.48 & 0.40 \\
 & & CNN & 1.75 & 1.02 & 2.49 & 0.00 \\
 &  & AJ & 1.30 & 1.03 & 1.56 & 0.00 \\
ME & \textit{israel} & BBC & 0.76 & 0.26 & 1.26 & 0.00 \\
&  & CNN & 0.49 & 0.09 & 0.89 & 0.02 \\
 &  & AJ & -1.93 & -2.96 & -0.91 & 0.00 \\
ME & \textit{palestine} & BBC & -0.31 & -1.39 & 0.78 & 0.58 \\
 &  & CNN & -0.49 & -1.45 & 0.46 & 0.31 \\
\bottomrule
\end{tabular}
    \caption{Log-Odds table with P-Values for tokens relevant to the analysis location: ``ukraine'', ``russia'',  and ``palestine'', ``israel'', and ``palestine''. We can observe that ``Russia'' is statistically significant for all source, with odds of appearing in the cause between 2.5 to 3 times more for cause then for effect.}
    \label{tab:logodds}
\end{table}

\newpage

\begin{figure}[h!]
    \centering
    \includegraphics[width=0.9\linewidth]{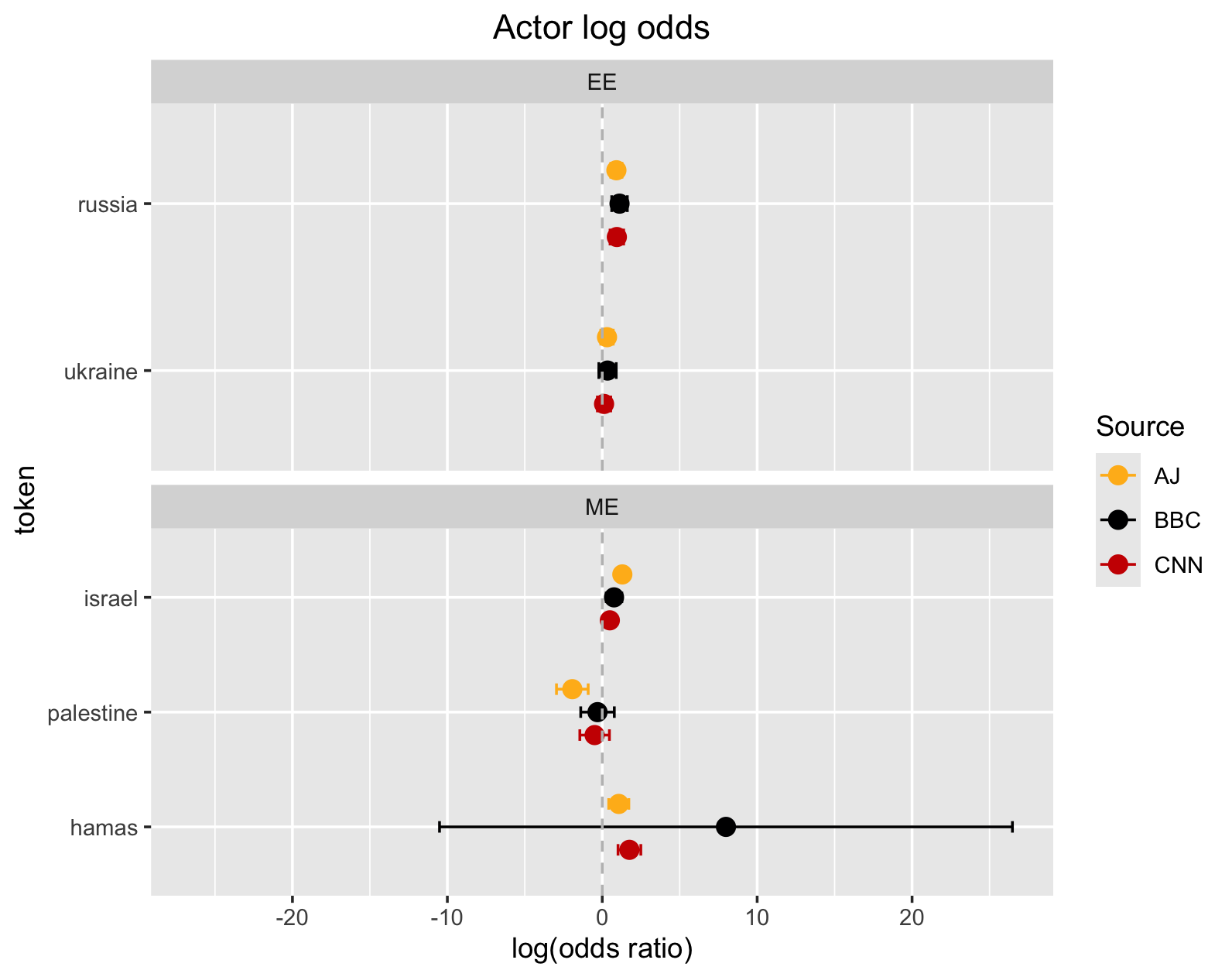}
    \caption{Reference to Figure \ref{fig:log-odds-actors} at correct scale.}
    \label{fig:appendix-actor-log-odds}
\end{figure}

\newpage

\subsection{Actor log-odds analysis on ``attack'' in headline subset}
\label{ssec:appendix_attack_in_headline}

\begin{table}[H]
    \centering
\begin{tabular}{lllrrrrr}
\toprule
location & tokens & source & log($\delta$) & CI lower & CI upper & p-value \\
\midrule
          & & AJ & 1.03 & 0.33 & 1.73 & 0.00 \\
EE & \textit{russia} & BBC & 0.68 & -0.41 & 1.77 & 0.22 \\
         & & CNN & 0.55 & -0.45 & 1.54 & 0.28 \\
         & & AJ & 0.67 & -0.08 & 1.43 & 0.08 \\
EE & \textit{ukraine} & BBC & 0.50 & -1.16 & 2.16 & 0.56 \\
        & & CNN & 1.13 & -0.34 & 2.61 & 0.13 \\
         & & AJ & 0.54 & -1.59 & 2.67 & 0.62 \\
ME & \textit{hamas} & BBC & 6.19 & -5.19 & 17.58 & 0.29 \\
        & & CNN & 1.54 & 0.09 & 2.98 & 0.04 \\
       &  & AJ & 2.14 & 1.25 & 3.04 & 0.00 \\
ME & \textit{israel} & BBC & 0.56 & -1.05 & 2.18 & 0.49 \\
        & & CNN & 0.12 & -0.93 & 1.17 & 0.82 \\
        & & AJ & -5.97 & -26.96 & 15.01 & 0.58 \\
ME & \textit{palestine} & BBC & 0.10 & -2.34 & 2.54 & 0.94 \\
        & & CNN & -0.20 & -2.76 & 2.35 & 0.88 \\
\bottomrule
\end{tabular}
    \caption{Statistical analysis of actors log-odds when ``attack'' is part of the headline. }
    \label{tab:attack-in-headline}
\end{table}

\end{document}